\documentclass[submission,copyright,creativecommons]{eptcs}
\providecommand{\event}{ICLP 2024} % Name of the event you are submitting to

\usepackage{iftex}
\usepackage{microtype}

\usepackage{inconsolata}
\usepackage{graphicx}
\usepackage{amsmath}
\usepackage{graphicx}
\usepackage{multirow}
\usepackage{times}
\usepackage{latexsym}
\usepackage{multirow}
\usepackage{algorithm}
\usepackage{algpseudocode}
\usepackage{hyperref}
\usepackage{amsmath,amssymb,trimclip,adjustbox}
\usepackage{verbatim}
\usepackage{mathtools}
\usepackage{booktabs}
\newcommand{\entails}{\mathop{:\!\!-}}

\usepackage[utf8]{inputenc}

\ifpdf
  \usepackage{underscore}         % Only needed if you use pdflatex.
  \usepackage[T1]{fontenc}        % Recommended with pdflatex
\else
  \usepackage{breakurl}           % Not needed if you use pdflatex only.
\fi

\title{Neuro-Symbolic Contrastive Learning for Cross-domain Inference}
\author{Mingyue Liu
\institute{University of Durham\\
United Kingdom}
\email{jflw25@durham.ac.uk}
\and
Ryo Ueda
\institute{University of Tokyo\\
Japan}
\email{ryoryoueda@is.s.u-tokyo.ac.jp}
\and
Zhen Wan
\institute{Kyoto University\\
Japan}
\email{zhenwan@nlp.ist.i.kyoto-u.ac.jp}
\and
Katsumi Inoue
\institute{National Institute of Informatics\\
Japan}
\email{inoue@nii.ac.jp}
\and
Chris G. Willcocks
\institute{University of Durham\\
United Kingdom}
\email{christopher.g.willcocks@durham.ac.uk}
}
\def\titlerunning{Neuro-Symbolic Contrastive Learning for Cross-domain Inference}
\def\authorrunning{M. Liu, R. Ueda, Z. Wan, K. Inoue \& C.G. Willcocks}

\begin{document}
\maketitle

\begin{abstract}
Pre-trained language models (PLMs) have made significant advances in natural language inference (NLI) tasks, however their sensitivity to textual perturbations and dependence on large datasets indicate an over-reliance on shallow heuristics. In contrast, inductive logic programming (ILP) excels at inferring logical relationships across diverse, sparse and limited datasets, but its discrete nature requires the inputs to be precisely specified, which limits their application. This paper proposes a bridge between the two approaches: neuro-symbolic contrastive learning. This allows for smooth and differentiable optimisation that improves logical accuracy across an otherwise discrete, noisy, and sparse topological space of logical functions.  We show that abstract logical relationships can be effectively embedded within a neuro-symbolic paradigm, by representing data as logic programs and sets of logic rules. The embedding space captures highly varied textual information with similar semantic logical relations, but can also separate similar textual relations that have dissimilar logical relations. Experimental results demonstrate that our approach significantly improves the inference capabilities of the models in terms of generalisation and reasoning.
\end{abstract}

\section{Introduction}

Deep neural network models have exhibited good precision in NLI tasks (\cite{nangia2019human,bowman2019deep}). However, the ability of these models to genuinely infer the logical relationship between sentences remains a topic of debate and controversy (\cite{gururangan2018annotation, sinha2020unnatural}). 
For example, it has been shown that labels can be detected solely by examining the hypothesis, without the need to examine the premise \cite{gururangan2018annotation}. Also, the model is incorrectly insensitive to the premise and hypothesis order; it should be sensitive to such shuffling \cite{sinha2020unnatural}. In addition, making inferences from simplified data pairs is challenging for the models that have been fine-tuned on MNLI or SNLI datasets \cite{luo2022simple}. The failure to learn the underlying generalisations raises doubts whether the models are relying on shallow heuristics to guess the correct label (\cite{mccoy2019right, luo2022simple, rosenman2020exposing, shen2022textual}).

In contrast to neural network models, Inductive Logic Programming (ILP), as a method of symbolic machine learning for reasoning tasks, can learn the relationships between input data and the target \cite{bratko1995applications}. The generalised logical rules can be induced from positive and negative examples in the form of predicate logic statements (\cite{muggleton1992inductive,cropper2022inductive}). %From the induced logic rules, it is also feasible to generate more data to fit it. 
The abstract data representation method makes ILP more data-efficient, generalised, and transferable for reasoning tasks. Also, logic-based programs tend to possess greater human interpretability, particularly when the predicates employed within the program represent concepts we are familiar with. %However, it is fragile when dealing with noise and mislabelled input data, and it cannot be directly applied on non-symbolic fields. 

\begin{figure*}
    \centering
\includegraphics[width=\textwidth]{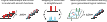}
    \caption{Logical data is discrete and sparse (red bars) and difficult to directly model (left blue curve) by a differentiable neural network $f_\theta$. However, we map meta-rules to-and-from the smooth PLM embedding space and utilise contrastive pairs (vertical arrows) to carve the sharp underlying logical structure (rightmost blue function) into $f_\theta$, enabling logical generalisation and logical reasoning.
    % It is difficult to directly fine-tune a deep neural network to be logically accurate due to sparsity and sudden changes in the topological space of logical functions. Straightforward to map continuous NL embeddings to discrete logical meta rules. }
    }
    \label{fig:pipeline1}
\end{figure*}

To combine the strength from both symbolic and connectionist sides (\cite{sen2022neuro, payani2019inductive}) and help neural language models to better capture the underlying logic structure, we propose a neuro-symbolic contrastive learning framework inspired by ILP, shown in Figure~\ref{fig:pipeline1}.  

In particular, we observe that the topological space of logical functions is difficult to accurately model with a PLM directly (Figure~\ref{fig:pipeline1}: left). Therefore we indirectly map from the natural language to the logical meta-rules (a relatively straightforward natural language task, Figure~\ref{fig:pipeline1}: centre). The meta-rules are assessed by the ILP to construct contrastive pairs that are used to fine-tune the PLM, ensuring dense representation of the underlying logical relationships (Figure~\ref{fig:pipeline1}: right), and thus improving overall PLM correctness and reasoning capability. This mapping process involves generating contrastive pairs that distinguish between logically consistent and inconsistent textual representations, thus carving a precise logical structure into the differentiable function of neural networks. The employment of hard examples—where positive pairs diverge lexically yet align logically, and negative pairs converge lexically but differ logically—facilitates a deeper engagement with the complexities of logical inference.

% In pursuit of this learning objective, we present a novel approach wherein we formulate and create triplet exemplar pairs. 
%This is therefore a neuro-symbolic data augmentation approach that constructs hard example pairs for model fine-tuning. Diverging from conventional techniques like retrieval-based methodologies that combine through training data, we construct symbolic NLI datasets. Leveraging the premise search algorithm and incorporating data augmentation techniques, we fabricate these NLI datasets from established ILP datasets. The part of the symbolic learning system enforces a consistent logic meta-rule for entailment across inference data, thereby signifying the profound inherent logical concordance within the reasoning process.

%Based on the augmented symbolic NLI datasets in predicate logic form, we build their corresponding datasets in natural language.
%To enhance the spectrum of textual representations, we employ the system LoLA, which is the extensive version of Grammatical Framework (\cite{calo2022enhancing, ranta2011grammatical}), to translate the datasets into natural language form.  A comprehensive array of rule templates is employed to ensure a broad diversity in the dataset. This methodology guarantees that the dataset encapsulates a wide variety of textual characteristics, including variations in sentence length and structural complexity.

Additionally, we enhance the symbolic NLI datasets, which are structured in predicate logic, by transforming them into their natural language equivalents employing the system of LoLA, an extension of the Grammatical Framework (\cite{calo2022enhancing}). This transformation leverages diverse rule templates to ensure a rich array of linguistic representations, effectively preparing the datasets to challenge the PLMs with a variety of textual and structural complexities. This approach to data augmentation ensures that our framework aligns with the practical demands of neuro-symbolic integration in natural language processing(NLP).

From Kautz’s Taxonomy, there are six levels of neuro-symbolic systems \cite{kautz2022third}. Our approach can be treated as a Level 3 \textsc{NEURO;SYMBOLIC} system, which is a hybrid framework whereby a neural network focusing on one task interacts with a symbolic system specialising in a complementary task. Our system utilises ILP for data augmentation tasks to construct hard example pairs to enhance the inference capabilities of neural networks.
%This research explores the integration of logic programming with neural network methodologies to advance reasoning capabilities in NLI tasks. 
The main contributions of this paper are:

%This research introduces novel integrations of logic programming techniques with neural network methodologies to enhance reasoning in Natural Language Inference (NLI) tasks. The contributions of this paper are meticulously designed to balance theoretical advancements and practical applications:

\begin{itemize}
    \item \textbf{Development of a Neuro-Symbolic Contrastive Learning Framework:} We introduce a framework that integrates Inductive Logic Programming (ILP) with the adaptive capabilities of contrastive learning in deep neural networks. This method enhances the logical reasoning abilities of neural models by utilising ILP-generated logical meta-rules to guide the training process, thus improving both performance and logical consistency. By differentiating between logically consistent and inconsistent textual representations through data augmentation of hard positive and negative example pairs, this framework effectively carves more precise underlying logical structures into the differentiable neural network function.

    \item \textbf{Transformation and Augmentation of Symbolic NLI Datasets:} Employing ILP, we develop symbolic NLI datasets that incorporate logical structures. These datasets are subsequently transformed into natural language using LoLA, an extension of the Grammatical Framework. The transformation process utilises diverse rule templates to ensure that the datasets exhibit comprehensive linguistic variability, which supports the practical application of these datasets in NLI tasks and demonstrates the application of logic programming principles in real-world scenarios.

    \item \textbf{Empirical Validation:} We assess the effectiveness of our neuro-symbolic framework against existing approaches under multiple settings. The analysis demonstrates improved performance in logical reasoning and generalisation, highlighting how the integration of logic programming can enhance the transferability of neural networks.

    \item \textbf{Theoretical Insights and Framework Implications:} Our research makes substantial theoretical contributions to the fields of logic programming and machine learning by exploring the potential of neuro-symbolic integration from the data augmentation aspect. We discuss the intuition of how this method can enhance the generalisability of the model.
\end{itemize}

\section{Background}

\subsection{Neuro-symbolic Frameworks for Reasoning}

The integration of neural networks with symbolic reasoning has given rise to neuro-symbolic frameworks, marking significant advancements in reasoning tasks and NLP. These frameworks aim to merge the adaptive capabilities of data-driven machine learning with the structured rigor of symbolic approaches, enhancing the complexity of linguistic analysis and understanding \cite{hamilton2022neuro}.

Recent studies by \cite{pendharkar2022asp} demonstrate the utility of Answer Set Programming (ASP) in encapsulating knowledge from natural language texts, providing a robust method for addressing complex queries directly from textual content. This method complements ASP-based approaches for declarative question answering, as further explored by \cite{mitra2019declarative}, which integrate external NLP modules to facilitate reasoning over natural language texts, thereby maintaining the contextual integrity of extensive texts. The integration of Meta-Interpretive Learning (MIL) with ASP, as detailed by \cite{kaminski2018exploiting}, illustrates how the incorporation of external sources can enhance the learning process by effectively managing the expansive search spaces encountered in MIL through efficient conflict propagation within the HEX-formalism.

The recent development of the Feed-Forward Neural-Symbolic Learner (FFNSL) underscores the potential of hybrid neuro-symbolic systems in deriving knowledge from raw data, such as images, by combining pre-trained neural models with logic-based machine learning systems to enhance both accuracy and interpretability \cite{cunnington2023ffnsl}. Furthermore, efforts by \cite{cunnington2022neuro} in Neuro-Symbolic Inductive Learning from raw data exemplify the integration of deep learning capabilities with symbolic reasoning to develop advanced AI systems capable of complex decision-making tasks.

Prominent models such as the Neural Logic Machine (NLM) employ probabilistic tensor representations to model logic predicates, simulating forward-chaining proof processes \cite{dong2019neural}. Similarly, the Differentiable Inductive Logic framework treats Inductive Logic Programming as a satisfiability problem, optimised through backpropagation \cite{evans2018learning,feng2020exploring}. Additionally, reinforcement learning has been utilised to create a neuro-symbolic framework that combines neural networks with natural logic, enhancing both elements \cite{tacl}. According to Kautz's Taxonomy, these approaches are categorised as Level 4 \textsc{Neuro:Symbolic → Neuro} systems, where symbolic rules are employed to direct neural training.

Other approaches, comparable to our own and categorised as Level 3 in Kautz’s Taxonomy, include the application of ILP to extract generalised logic rules from Knowledge Graphs (KG), which utilise advanced search algorithms and pruning techniques \cite{zhang2021neural}. The Neuro-Symbolic Concept Learner (NS-CL), for example, captures visual concepts and linguistic terms to construct scene representations grounded in symbolic programs \cite{mao2019neuro}. Furthermore, DeepProbLog integrates symbolic reasoning with neural perception to solve tasks that require both high-level and low-level cognitive processes \cite{manhaeve2018deepproblog}.

\subsection{Preliminary of Inductive Logic Programming}
% As a subfield of symbolic machine learning, Inductive Logic Programming (ILP) induces a set of logical rules (clauses) that generalizes training examples. Whereas most forms of machine learning learn functions, ILP learns relations \cite{muggleton1994inductive,cropper2020inductive}. ILP mainly focuses on learning Horn clause — clauses with at most one positive literal, as the following form:
%修
As a subfield of symbolic machine learning, \emph{Inductive Logic Programming} (ILP) induces a set of logical rules (clauses) that generalises training examples.
%Unlike most forms of machine learning that learn functions,
ILP learns relations rather than functions \cite{muggleton1994inductive,cropper2022inductive}.
ILP mainly focuses on learning Horn clause --- clause with at most one positive literal, as the following form:
\begin{equation}
    h \entails b_1, b_2, \dots, b_n,
\end{equation}
which stands for the implicational form:
\begin{equation}
    \mathrm{h} \leftarrow \mathrm{b}_1 \wedge \mathrm{b}_2 \wedge \dots \wedge \mathrm{b}_n.
\end{equation}
% A clause with this form states that, if all the conjuncted \textsc{Body} atoms are true, the \textsc{Head} atom is true. Every \textsc{atom} is a formula $p(t_1, t_2,..., t_n)$, where $p$ is a \textsc{predicate} symbol of arity $n$ and each $t$ is a term (constant or variable). A predicate is an $n$-ary Boolean function states the relations of terms. 
%修改版本
This is a Horn clause, meaning that, if all the conjuncted \emph{Body} atoms $b_{1},\ldots,b_{n}$ are true, then the \textit{Head} atom $h$ is true.
Every \emph{atom} is a formula $p(t_1, t_2,..., t_n)$, where $t_i$ is a term (a constant or a variable) and $p$ is a \emph{predicate} symbol of arity $n$.
%, an $n$-ary Boolean function states the relations of terms. 

% A clausal theory $T$ is a set of clauses. When clause $C$ is the consequence of theory $T$, C is the entailment from $T$, denoted by $T\models C$. The learning goal of ILP is obtaining a set of hypothesis $H$, which is the assumed relationships induced from background knowledge $B$. From the set of background knowledges $B$, the sets of positive examples $E^+$ and negative examples $E^-$ can be obtained from $B$ based on closed world assumption \cite{reiter1981closed}.

%修改版本
A \emph{clausal theory}, denoted as $T$, is a collection of clauses.
If a clause $C$ is a consequence of the theory $T$, then $C$ is the entailment from $T$, denoted as $T\models C$.
The learning objective of ILP is obtaining an \emph{explanation} $H$, which is the assumed relationship induced from \emph{background knowledge} $B$. In ILP, positive examples $K^+$ and negative examples $K^-$ are given as input.
In logical words, this is
\begin{equation*}
    \begin{cases}
    \forall k \in K^+, H \cup B \vDash k \quad \text{($H$ is complete)}, \\
    \forall k \in K^-, H \cup B \nvDash k \quad \text{($H$ is consistent).}
    \end{cases}
\end{equation*}
% The learning of entailment aims to learn a set of $H$:
% \begin{equation}
%     \begin{cases}
%     \forall e \in E^+, H \bigcap B \vDash e 
%     (i.e. H ~is~ Complete)\\
%     \forall e \in E^-, H \bigcap B \nvDash e 
%     (i.e. H ~is~ Consistent).
%     \end{cases}
% \end{equation}

% The Herbrand interpretation set $I$ is a Herbrand model \cite{cropper2022inductive} for a set of clauses $C$ when
% \begin{equation}
%     \begin{cases}
%             h_1; h_2; \dots; h_m  \coloneq  b_1,b_2,\dots,b_n\in C,  \\
%     \begin{split}
%     \theta\colon &\{ b_1\theta, b_2\theta,\dots, b_n\theta \}\subset I \rightarrow \\
%     &\{ h_1\theta, h_2\theta,\dots, h_m\theta \}\cap I \neq \varnothing,
%     \end{split}
%     \end{cases}
% \end{equation}
% where $\theta$ stands for substitution function for $V_n$ variables.
%修改版本
A Herbrand interpretation $I$ is a subset of the Herbrand base, and is a Herbrand model of a set $T$ of clauses $C$ when
\begin{equation*}
    \begin{cases}
            \text{For each }  (h  \entails  b_1,b_2,\dots,b_n)\in T,  \\
    %\begin{split}
    \text{if }\exists\theta\colon \{ b_1\theta, b_2\theta,\dots, b_n\theta \}\subset I, \text{ then } 
    h\theta \in I.
%    \end{split}
    \end{cases}
\end{equation*}
 $\theta=\{ v_1/t_1, \dots, v_n/t_n \}$ is a substitution function which replaces variables $\{ v_1, \dots, v_n\}$ in a clause with terms $\{ t_1, \dots, t_n\}$.

\subsection{Introduction of NLI Task}
Natural Language Inference (NLI) is a fundamental task in computational linguistics where a system is tasked with determining the logical relationship between a pair of sentences, known as the \textit{premise} and the \textit{hypothesis}\footnote{Please do not confuse this notion of hypotheses for NLP with
those hypotheses in ILP.  The mainstream benchmarks and datasets in NLP community call it \textit{hypothesis} \cite{gao2021simcse,shen2022textual,williams2017broad}. We thus have two kinds of hypotheses with different notations and meanings for ILP and NLP.  In this paper, we call a \textit{hypothesis} $H$ in ILP as an \textit{explanation} $H$ (Section
2.2).}. Specifically, the goal is to ascertain whether the \textit{hypothesis} is true (entailment), false (contradiction), or indeterminate (neutral) based on the information in the \textit{premise} \cite{bos2005recognising,bowman-etal-2015-large}. This task mimics key aspects of human reasoning and is crucial for testing the ability of systems to perform logical inference.

NLI is pivotal for advancing AI technologies that necessitate a nuanced comprehension of natural language. It challenges computational models to interpret subtleties inherent in human communication, such as ambiguity, contextual implications, and inferential logic \cite{bowman2019deep,williams2017broad}. We aim to enhance the interpretability and reliability of models through the integration of logic programming within Natural Language Inference (NLI) research, thereby advancing the capabilities of machines to process and interact with human language in a logically coherent manner.

\subsection{Contrastive Learning for NLI}
Contrastive learning is a machine learning technique that enhances the discriminative capabilities of models by enabling them to differentiate features between similar and dissimilar data instances. Originally prominent in computer vision, this technique has been effectively adapted for natural language processing (NLP), where it is used to refine a model's ability to parse and understand complex textual relationships. In the NLP domain, models are trained using pairs of data instances—positive pairs, which are semantically similar, and negative pairs, which are semantically dissimilar—thereby training the model to recognise subtle textual nuances \cite{chen2020simple,9157636}.

The use of Natural Language Inference (NLI) datasets, such as SNLI \cite{bowman-etal-2015-large} and MultiNLI \cite{williams-etal-2018-broad}, has been instrumental in providing supervised annotations for contrastive learning. Techniques like Supervised SimCSE leverage entailment pairs as positive examples and use contradiction pairs and other unrelated in-batch instances as negative examples to fine-tune models' semantic understanding \cite{gao2021simcse}. SBERT, employing a siamese architecture with a shared BERT encoder, further illustrates the application of these datasets to train on discerning semantic discrepancies \cite{reimers-gurevych-2019-sentence}. Additionally, self-supervised approaches often utilise methods such as back translation, dropout, and token shuffling to create contrastive learning pairs, enhancing the model's robustness by exposing it to a diverse array of linguistic transformations \cite{gao2021simcse,yan2021consert}.

Hard examples, or those data pairs that are challenging for the model to correctly classify due to their nuanced differences or similarities, are particularly crucial in the training process of contrastive learning \cite{le2020contrastive,oh2016deep}. These examples help in refining the model's ability to perform fine-grained distinctions and to generalise better to unseen data. In contrastive learning, hard positive pairs may include sentences with substantial lexical divergence yet sharing a similar meaning, whereas hard negative pairs might consist of sentences that are lexically similar but diverge in meaning \cite{ shen2022textual}. Generating these challenging pairs requires sophisticated data augmentation techniques that can manipulate textual and logical features effectively.

Our proposed method emphasises the creation and utilisation of such hard examples by identifying positive pairs that exhibit textual differences yet share logical similarities, and negative pairs that appear similar but differ in logic. This focus is implemented through an advanced hybrid framework that combines symbolic reasoning with neural processing, aiming to enhance the model's deep linguistic and logical understanding, which is essential for complex tasks like NLI.

\subsection{Problem formulation}

%Contrastive Learning (CL) in neural networks involves learning to distinguish between closely related concepts by contrasting positive examples with negative examples. It uses an anchor data point as a reference to measure similarity or dissimilarity.

In the context of our neuro-symbolic CL framework, the traditional logical terms are adapted with specific meanings:
\begin{itemize}

    \item \textbf{Anchor data point} (\( E \)): In our framework, an anchor data point \( E \) consists of a pair \( (P, L) \), where \( P \) is the premise and \( L \) is the conclusion derived from \( P \). The anchor serves as the reference point for comparison against other examples in the dataset.
    \item \textbf{Premise} (\( P \)): A statement or proposition that provides the context from which the conclusion \( L \) is logically inferred.
    \item \textbf{Hypothesis} (\( L \)): A logical conclusion that consistently follows from the premise. Instead of the term `conclusion', in the standard CL and NLI setups \cite{gao2021simcse,shen2022textual,williams2017broad}, they previously termed as `hypothesis' here. \( L \) can be labelled as true (entailment), false (contradiction), or indeterminate (neutral).

    \item \textbf{Hard Positive Examples} (\( E^+ \)): Composed of \( (P^+, L^+) \), where \( P^+ \) and \( L^+ \) adhere to the same logical rule as \( P \) and \( L \) but vary in textual or domain characteristics. This setup ensures that \( L^+ \) is a valid conclusion under the same premises but presented differently. The \( L^+ \) means it is a hard positive example relative to \( L \), not an indication of \( L \)'s truth value.
    
    \item \textbf{Hard Negative Examples} (\( E^- \)): Constructed as \( (P^-, L^-) \), these examples share textual similarity with \( P \) but lead to \( L^- \), a conclusion that logically contradicts or deviates from \( L \) under the given premise. The \( L^- \) represents a hard negative example relative to \( L \), challenging the model's ability to discern subtle logical distinctions and is not a label of \( L \) being false.

\end{itemize}

The primary objectives of our contrastive learning framework are formally defined as follows:

\[
\begin{cases}
\mathop{\textrm{minimize}} \, d(E, E^+) & : \text{ to enforce logical consistency}, \\
\mathop{\textrm{maximize}} \, d(E, E^-) & : \text{ to capitalise on logical deviations},
\end{cases}
\]where \( d \) denotes a distance function (metric) in the embedding space. The minimisation objective aims to align embeddings of \( E \) and \( E^+ \), which are logically consistent. Conversely, the maximisation objective aims to differentiate between embeddings of \( E \) and \( E^- \), which represent logical deviations, thereby enhancing the model's ability to discern fine-grained logical distinctions.

\section{Methodology}\label{methodology}

Inspired by ILP, we construct symbolic NLI datasets by augmentation that maximises textual variability while maintaining logical consistency. Every augmented dataset consists of two subsets represented as predicate logic forms and natural language forms. 

\begin{figure}
    \centering
    \includegraphics[width=0.4\textwidth]{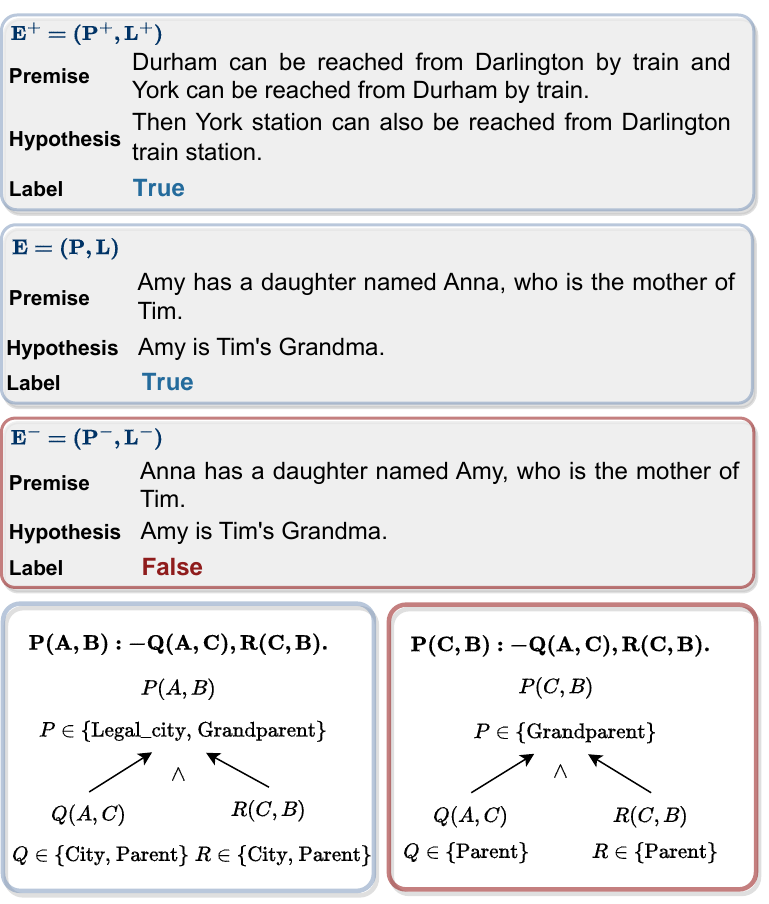}
    \caption{Illustration of an anchor data point $E = (P,L)$ with its corresponding positive and negative pairs. The positive pair $E^+ = (P^+, L^+)$ maintains logical consistency with the anchor, while the negative pair $E^- = (P^-, L^-)$ introduces a logical contradiction despite overlapping textual content.}
    \label{fig1}
\end{figure}

For the logical form, we use symbolic learning systems to enforce a consistent meta-rule for conclusions across inference data, which indicates the high underlying logical similarity of reasoning process. 
And for the natural language, we translated from the corresponding logic form via Grammatical Framework (GF) with various rule templates to ensure diversity in textual representations, such as length and complexity. 
Moreover, we propose an ILP-inspired Contrastive Learning framework to further boost the performance of models on cross-domain inference tasks. For each anchor data point $E = (P, L)$, where $P$ is the premise and $L$ is the hypothesis (conclusion), we construct hard positive example pairs $E^+ = (P^+, L^+)$, which share the same logic meta-rule but originate from different textual domains. Conversely, a hard negative example pair consists of an anchor point and a hard negative data point $E^- = (P^-, L^-)$ within the same domain, which is textually similar but logically different.

 As shown in Figure~\ref{fig1}, given an anchor data point denoted as $E = (P, L)$ (where $P$ signifies the premise and $L$ represents the hypothesis), we generate hard positive example pairs $E^+ = (P^+, L^+)$. The hard positive example pairs share an identical logic meta-rule yet originate from distinct domains. Conversely, the formulation of a hard negative example pair involves an anchor point and a challenging negative data point $E^- = (P^-, L^-)$ within the same domain. This pair exhibits textual similarity while diverging logically.

The first two examples shown in blue colour are varying in domains and textual representation, while the red-coloured example has high token-level overlapping with the middle case. However, the logic rules below these three examples indicate that the underlying logic meta-rule of the low token-level overlapping examples are identical, while the higher textual similarity ones are logically different.

  Our method seeks to learn an embedding space in which the vector representations of $E$ and $E^+$  are close together, due to the fact that they share the same mathematical logic reasoning process to inference, despite the difference in their textual expression and domains. On the other hand, since $E$ and $E^-$ have similar textual expressions but divergent mathematical logical reasoning processes, their vector representations should be separated.

We will explain the details of each part of our methodology in the following sections.
\subsection{Meaning Representations and Dataset Construction}
% A standard ILP dataset is formed of three sets of components: Background Knowledge ($B$), Positive Examples ($K^+$) and Negative Examples ($K^-$). As we introduced in section 1.2, ILP aims to induce a set of hypothesis rules that with the $B$ entails $k \in K^+$ and contradicts $k \in K^-$ \cite{cropper2020inductive}. Here is a toy example of one of the ILP datasets we used:
% 改
A standard Inductive Logic Programming dataset is formed of three sets of components: background knowledge ($B$), positive examples ($K^+$), and negative examples ($K^-$). As we introduced in section 1.2, ILP aims to induce a set of rules that with the $B$ entails $k \in K^+$ and contradicts $k \in K^-$ \cite{cropper2022inductive}. The following is a toy example of one of the ILP datasets we used:
\begin{gather*}
    B=
    \begin{cases}
        \mathrm{parent}(\mathrm{Ann}, \mathrm{Amy})    \\
        \mathrm{parent}(\mathrm{Amy}, \mathrm{Amelia}) \\
        \mathrm{parent}(\mathrm{Amy}, \mathrm{Andy})   \\
        \mathrm{parent}(\mathrm{Linda}, \mathrm{Garin})
    \end{cases}\\
    K^+=
    \begin{cases}
        \mathrm{grandparent}(\mathrm{Ann}, \mathrm{Amelia}) \\
        \mathrm{grandparent}(\mathrm{Linda}, \mathrm{Amelia})
    \end{cases}\\
    K^-=
    \begin{cases}
        \mathrm{grandparent}(\mathrm{Amy}, \mathrm{Amelia}) \\
        \mathrm{grandparent}(\mathrm{Amelia}, \mathrm{Ann})
    \end{cases}
\end{gather*}
        % \begin{gather*}
        %     B=
        %     \begin{cases}
        %         parent(Ann, Amy)    \\
        %         parent(Amy, Amelia) \\
        %         parent(Amy, Andy)   \\
        %         parent(Linda, Garin)
        %     \end{cases}\\
        %     K^+=
        %     \begin{cases}
        %         grandparent(Ann, Amelia) \\
        %         grandparent(Linda, Amelia)
        %     \end{cases}\\
        %     K^-=
        %     \begin{cases}
        %         grandparent(Amy, Amelia) \\
        %         grandparent(Amelia, Ann)
        %     \end{cases}
        % \end{gather*}
Every positive/negative examples is matched with the corresponding necessary premise from $B$. The following Algorithm~\ref{alg:nodesearch} shows the search algorithm for the premise filtering process.

\begin{algorithm}
\caption{Premise Search}
\label{alg:nodesearch}
\small
\textbf{Input:} $B$, $R$ (set of $t$ for every $k \in K^+/K^-$) \\
\textbf{Parameter:} Optional list of parameters\\
\textbf{Output:} $\mathrm{filtered\_premise\_list}$

\begin{algorithmic}[1] % Adding [1] enables line numbering
\For{$\mathrm{predicate} \textbf{ in } B$}
    \If{$\mathrm{predicate}.t \textbf{ in } R$}
        \State $\mathrm{filtered\_premise\_list}.\textbf{insert}(\mathrm{predicate})$
        \If{$\mathrm{predicate}.t.\mathrm{rest} \textbf{ not in } R$}
            \State $R.\textbf{insert}(\mathrm{predicate}.t.\mathrm{rest})$
        \EndIf
    \EndIf
\EndFor
\end{algorithmic}
\end{algorithm}
%一段伪代码

Hence, the logic rules extracted from the toy example is given by
%Formula
% \begin{align}
% \begin{split}
% \footnotesize
%                parent(Ann, Amy) \wedge parent(Amy, Amelia) & \\ \rightarrow  grandparent(Ann, Amelia) \quad \text{[Positive]}&, 
% \end{split}
%  \\
%  \begin{split}
%         parent(Ann, Amy)  \wedge parent(Amy, Amelia)& \\  \rightarrow grandparent(Amelia, Ann) \quad \text{[Negative]}&,\end{split}
% \end{align}
%\begin{equation}
\begin{equation}
    \mathrm{grandparent}(\mathrm{Ann}, \mathrm{Amelia}) \entails \mathrm{parent}(\mathrm{Ann}, \mathrm{Amy}), \mathrm{parent}(\mathrm{Amy}, \mathrm{Amelia}),
\end{equation}

\begin{equation}
    \mathrm{grandparent}(\mathrm{Amelia}, \mathrm{Ann}) \entails \mathrm{parent}(\mathrm{Ann}, \mathrm{Amy}), \mathrm{parent}(\mathrm{Amy}, \mathrm{Amelia}).
\end{equation}

%\end{equation}
And the constructed NLI dataset is shown in Table~\ref{table1}, where predicates $p$ and $gp$ stand for parent and grandparent respectively.
% \begin{figure}[h]
% 	\centering
% 	\includegraphics[width=0.47\textwidth]{fig/List.png}
%         \caption{Toy examples.}
%   \label{fig_3}
% \end{figure}

\begin{table}[ht]
%\small
\centering
\caption{Toy examples of the constructed NLI dataset, where `$+$', `$-$', and `$\mathrm{N}$' labels denote true
(entailment), false (contradiction), and indeterminate (neutral) respectively.}
\renewcommand{\arraystretch}{0.67} 
\label{table1}
\begin{tabular}{llc}
\hline
\textbf{Premise} & \textbf{Hypothesis} & \textbf{Label} \\
\hline
$p(\text{Ann}, \text{Amy}), p(\text{Amy}, \text{Rita})$ & $gp(\text{Ann}, \text{Rita})$ & $+$ \\
$p(\text{Ann}, \text{Amy}), p(\text{Amy}, \text{Rita})$ & $gp(\text{Rita}, \text{Ann})$ & $-$ \\
$p(\text{Ann}, \text{Amy}), p(\text{Amy}, \text{Rita})$ & $gp(\text{Linda}, \text{Garin})$ & $\mathrm{N}$ \\
\hline
\end{tabular}
\end{table}

\begin{table*}[ht]
\centering
%\small
\renewcommand{\arraystretch}{0.67} 
\caption{Augmented samples from the given toy examples in Table~\ref{table1}.}
\label{datasetaug}
\begin{tabular}{llc}
\hline
\textbf{Premise} & \textbf{Hypothesis} & \textbf{Label} \\
\hline
$p(\text{Amy}, \text{Amelia}), p(\text{Ann}, \text{Amy}), p(\text{Amy}, \text{Andy})$ & $gp(\text{Ann}, \text{Amelia})$ & $+$ \\
$p(\text{Alex}, \text{Joe}), p(\text{Joe}, \text{Charles})$ & $gp(\text{Charles}, \text{Alex})$ & $-$ \\
$p(\text{Joe}, \text{Charles}), p(\text{Alex}, \text{Joe}), p(\text{Amy}, \text{Amelia}), p(\text{Linda}, \text{Garin})$ & $gp(\text{Charles}, \text{Linda})$ & $\mathrm{N}$ \\
\hline
\end{tabular}
\end{table*}
We systematically augment datasets using a variety of methods tailored to maintain logical integrity while introducing structural variability. These methods include constructing templates for replacing constants in the terms $t_i$ of predicates $p(t_1, t_2, ..., t_n)$, appending logically irrelevant predicates to the premises, and permuting the order of premise predicates to demonstrate the invariance of logical conjunctions under operand permutation. For example, in a toy dataset, the predicates within a premise can be reordered or terms $t_i$ substituted using an alternative lexicon to test the robustness of logical inference models to syntactic variations. Table \ref{datasetaug} lists some possible sample data after augmentation.

\subsection{Metarules of Cross-domain Tasks}

%Metarules are a popular form of syntactic bias and are used by a range of ILP systems \cite{cropper2022inductive}. 
Different from the usual usage of metarules \cite{muggleton2012ilp, cropper2022inductive}, we apply metarules here to construct hard positive examples for contrastive learning. %Metarules are second-order rules which define the logical skeleton of logic formula, which in turn defines the hypothesis space \cite{cropper2020inductive}. 
As shown in Figure \ref{fig1}, $E^+ = (P^+,L^+)$ and the anchor data point $E = (P, L)$ share the same metarule below: %to learn the grandparent relation given the parent relation, the chain metarule would be suitable:
\begin{equation}
   P(A,B) \entails  Q(A,C), R(C,B).
   \label{rule1}
\end{equation} 
First-order variables are denoted by the letters $A$, $B$, and $C$, whereas second-order variables are denoted by the letters $P$, $Q$, and $R$. %First-order variables can be substituted by constant symbols, and second-order variables can be substituted by predicate symbols. For instance, as shown in Figure \ref{fig1}, to represent the relationship between the grandparents given the parents,
The substitution functions of the second-order variables $P$, $Q$, and $R$ are
% \begin{equation}
% \mathrm{substitutions} \{ P/\mathrm{legalCity}, Q,R/\mathrm{city} \},
% \end{equation}
% \begin{equation}
%      \mathrm{substitutions} \{ P/\mathrm{grandparent}, Q,R/parent \}.
%      \label{sub1}
% \end{equation} 
\begin{align}
&\mathrm{substitutions} \{ P/\mathrm{legalCity}, Q,R/\mathrm{city} \}, \\
&\mathrm{substitutions} \{ P/\mathrm{grandparent}, Q,R/\mathrm{parent} \}. \label{sub1}
\end{align}
After applying the substitution functions, the induced logical relationship between $\mathrm{parent}$ and $\mathrm{grandparent}$ ($gp$) is

\begin{equation}
    \mathrm{gp}(A, B) \entails \mathrm{parent}(A, C), \mathrm{parent}(C, B),
    \label{equa1}
\end{equation}
and the transition logic rule of accessible transportation between cities is
\begin{equation}
    \mathrm{legalCity}(A, B) \entails \mathrm{city}(A, C), \mathrm{city}(C, B).
    \label{equa2}
\end{equation}

Logic rules (\ref{equa1}) and (\ref{equa2}) are isomorphic since they share the same metarule and there exists a bijective substitution function $\theta$ to make them logically equivalent.%\cite{amgoud2018measuring}. 

On the other hand, as shown in Figure \ref{fig1}, although $E^- = (P^-,L^-)$ and the anchor data point $E = (P, L)$ are textually similar and from the same domain of $\mathrm{parent}$, $E^- = (P^-,L^-)$ has a different metarule from rule (\ref{rule1}) 
\begin{equation}
   P(C,B) \entails  Q(A,C), R(C,B).
\end{equation} 
And it cannot be logically equivalent with the rule (\ref{equa1}) after applying substitution function (\ref{sub1}).

\subsection{Data Augmentation for Contrastive Learning}

For each anchor data point \( E=(P,L) \), we construct its hard positive data point \( E^+=(P^+,L^+) \) and hard negative data point \( E^-=(P^-,L^-) \). The premise \( P \) is represented as a conjunction of body predicates \( b \), where \( P = \{b_1, b_2, \dots, b_n \} \). The contrastive learning approach uses the \( \mathcal{L}_{cl} \) loss to pull the representation of \( E \) closer to \( E^+ \) and push it away from \( E^- \), which sharpens the model’s ability to discriminate between subtle variations in logical coherence.

Through a permutation step defined by $\sigma$, we reorder $b$ to obtain $b'= p(t_{\sigma(1)}, t_{\sigma(2)}, \dots, t_{\sigma(n)})$. This permutation introduces variability in the data structure, aiding the model in learning to recognise essential logical constructs regardless of their syntactic presentation.

\subsection{Hard Positive Example Pairs}

 In the scenario of a hard positive example pair, $E$ and $E^+$ are connected by a substitution function $\theta=\{ v_1/t_1, \dots, v_n/t_n \}$, aligning them under the condition $E\theta = E^+\theta$. Notably, the variables \{$v_1, \dots, v_n\} \in \mathcal{D}_1$ and the terms $\{t_1, \dots, t_n \} \in \mathcal{D}_2$, where $\mathcal{D}_1$ and $\mathcal{D}_2$ signify distinct domains.

\subsection{Hard Negative Example Pairs}
Given a premise $P = \{ b_1, b_2, \dots, b_n\}$ and an hypothesis (conclusion) $L = \{h\} = \{ p(t_1, t_2, \dots, t_n)\} $, we choose an arbitrary $b_i \in P$ such that $b_i = p_i(t_1, t_2, \dots, t_n)$. 

One way of constructing a hard negative example is permuting $b_i$ to obtain $E^-_1 = (P^-, L^-)$ with $P^-=\{b_1,\dots,b'_i,\dots, b_n\}$. Another way is permuting $L$ to get $E^-_2=(P^-,L^-)$ with $L^-=\{ h' \}$. 

\subsection{Training Process of Contrastive Learning}

Contrastive learning will be performed on triplets pairs $(E_i,E^+,E^-)$. The training objective $(x_i,x^+,x^-)$ with batch size $N$ is
\begin{equation}\label{clloss}
%\small
\mathcal{L}_{\mathrm{cl}} = -\mathbb{E} \left[ \log \frac{e^{\cos{(x_i, x_i^+)}/\tau}}{\sum_{j=1}^{N}\!\left(e^{\cos{(x_j, x_j^+)}/\tau\!} + e^{\cos{(x_j, x_j^-)}/\tau\!}\right)\!} \right],
\end{equation}
where $x_i$ denotes the encoder representation of $E_i$ (\cite{gao2021simcse}). The \( \mathcal{L}_{cl} \) loss function employs cosine similarity in the embedding space to evaluate the closeness of embeddings. The encoder used for generating representations \( x_i \) is typically a neural network such as a Transformer or LSTM \cite{vaswani2017attention, hochreiter1997long}. These architectures are chosen due to their proficiency in capturing contextual relationships in text, crucial for the nuanced understanding required in NLI tasks.

\subsection{Rule-based Translation between Logic-form and Natural Language }
\begin{figure}
    \centering
\includegraphics[width=0.9\textwidth]{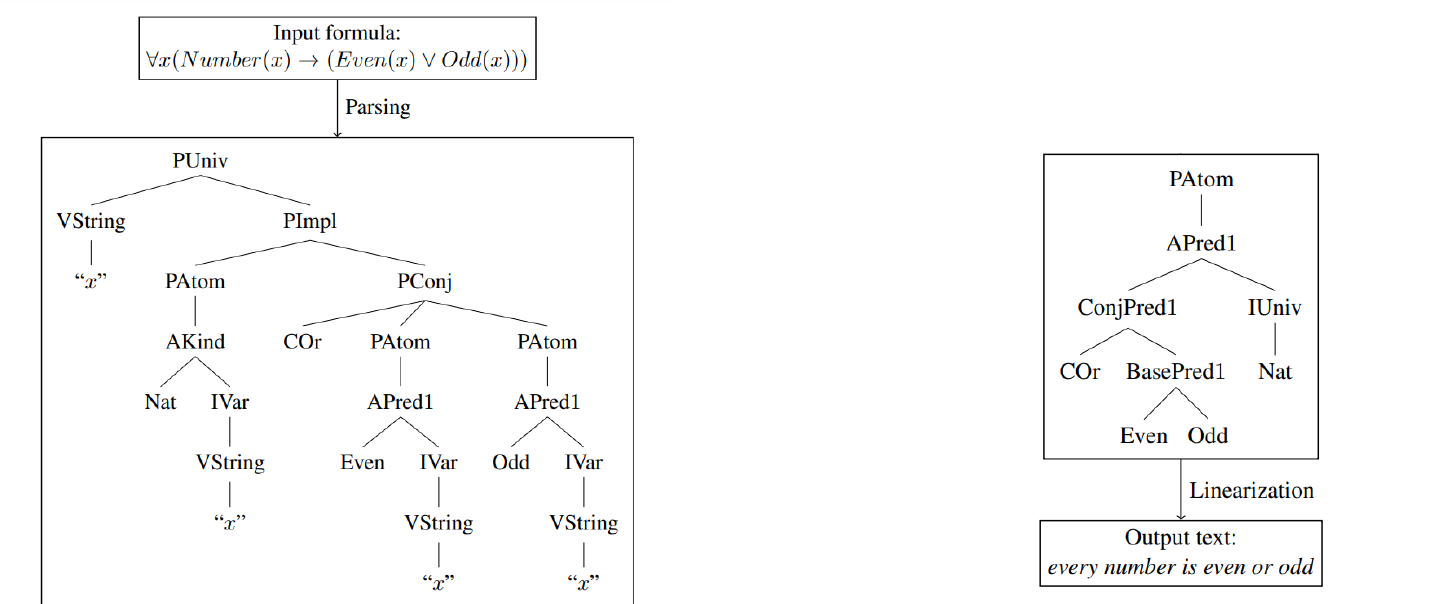}
    \caption{A model of the translation system is presented, including an example of translating a First-Order Logic (FOL) formula into English. Each node in the Abstract Syntax Tree (AST) is named after the syntactic function used to construct the corresponding constituent \cite{calo2022enhancing}. The right side of this figure displays the tree structure following an optimisation step applied to the initial configuration on the left side.
    }
    \label{lola}
\end{figure}

We use LoLA (\cite{calo2022enhancing}), which is the extensive version based on Grammatical Framework (GF) (\cite{ranta2004grammatical}) to enable the translation between natural language and propositional logic formulas. The translation is purely rule-based. Initially, the expression in the source language undergoes parsing, resulting in the derivation of an abstract syntax tree (AST). Subsequently, the AST undergoes a linearisation process, yielding a linguistic manifestation in the target language through the utilisation of language-specific concrete syntax conventions (\cite{calo2022enhancing}). Figure~\ref{lola} shows the toy example of the translation system. To make the translated natural language more understandable, for input logical formulas, LoLA uses logical equivalence laws to search for the optimal expression and remove redundant information. 

%As discussed in the section on ILP, each set of premises $P$ and their corresponding hypothesis(conclusion) $L$ constitute a Horn clause. A Horn clause in its implication form is logically equivalent to its disjunctive normal form. Thus, the following two logical expressions are equivalent, while their textual representations differ:

%\begin{equation}
%    \mathrm{parent}(A,C) \wedge \mathrm{parent}(C,B) \rightarrow \mathrm{gp}(A,B).
%    \label{equa33}
%\end{equation}
%\begin{equation}
%    \neg \mathrm{parent}(A,C) \vee \neg \mathrm{parent}(C,B) \vee \mathrm{gp}(A,B).
%    \label{equa44}
%\end{equation}
To enhance the comprehensibility of natural language translations derived from logical formulas, we utilise logical equivalence laws to generate varied yet equivalent expressions. The NLI dataset, constructed from these equivalent but textually distinct forms, ensures consistent truth labelling, which is crucial for the construction of hard examples. We constructed various rule templates to enable the generation of more diverse datasets, varying in textual length and reasoning difficulty. Here are some examples shown in Figure~\ref {fig1}, Table \ref{tableemotion} and Table~\ref{tab:logicexamples}:

\begin{table}[H]
\centering
\caption{Examples of equivalent transformations where $\mathrm{par}$ and $\mathrm{gp}$ denote parent and grandparent respectively.}
\renewcommand{\arraystretch}{0.97} 
\label{tableemotion}
\begin{tabular}{lcc}
    \hline
    \textbf{Premise} & \textbf{Hypothesis} & \textbf{Label} \\ 
    \hline
    $\mathrm{par}(A,C) \wedge \mathrm{par}(C,B)$ & $\mathrm{gp}(A,B)$ & $+$ \\ 
    $\neg \mathrm{par}(A,C) \vee \neg \mathrm{par}(C,B) \vee \mathrm{gp}(A,B)$ & $\mathrm{gp}(A,B)$ & $+$\\
    \hline
\end{tabular}
\end{table}

\begin{table}[htbp]
\centering
\caption{Examples of Logic Rules and Corresponding translated Natural Language Premises and Hypotheses.}
\label{tab:logicexamples}
\begin{tabular}{p{0.95\textwidth}}
\toprule
\textbf{Logic Rule:} $\text{legalCity(Delwino, Borovan)} \entails \text{City(Delwino, Ebadong)} , \text{City(Ebadong, Borovan)}$ \\
\textbf{Premise:} From Delwino, one can take a train to Ebadong. And from there, it is possible to travel to Borovan by train. \\
\textbf{Hypothesis:} Therefore, the train network connects Delwino and Borovan. \\
\textbf{Label:} Entailment \\
\midrule
\textbf{Logic Rule:} $\text{legalCity(Guinimanan, Ersama)} \entails \text{City(Jenau, Ersama)} , \text{City(Kotla Pehluan, Ersama)}$, \\
$\text{City(Jalawanan, Sangbanwol)}$ \\
\textbf{Premise:} The city Ersama can be accessed by bike from Jenau. Sangbanwol is connected to Jalawanan by train, and you can take a train from Ersama to Kotla Pehlwan. \\
\textbf{Hypothesis:} These will allow you to reach Guinimanan from Ersama. \\
\textbf{Label:} Neutral \\
\bottomrule
\end{tabular}
\end{table}

%\begin{figure}[tbh!]
%	\centering
%	\includegraphics[width=0.56\textwidth]{fig/translate.png}
%	\caption{The examples of translation.}
%        \label{translate}
%\end{figure}

\section{Experiment and Result}
\subsection{Dataset}
We select some of the classic ILP task datasets --- the ancestor dataset from GILPS (General Inductive Logic Programming System) and the kinship dataset from Popper \cite{cropper2021learning}. Every dataset is built with three components all in predicate logic arguments: Background Knowledge ($B$), Positive Examples ($K^+$), and Negative Examples ($K^-$). 

\textbf{[City Transportation Dataset]} is a self-proposed dataset with $B$ of train connections between two cities and $K^+$ and $K^-$ represent feasible transportation between cities.

\textbf{[Popper: Kinship Dataset]} is a minimal ILP dataset for kinships. $B$ gives parent relationships and $K^+$ and $K^-$ give examples for grandparent relationships.

\textbf{[GILPS: Ancestor Dataset]} is an ILP dataset for relationships between a big family tree. $B$ provides information on gender, names, and parent relationships between every generation. And $K^+$ and $K^-$ are examples of ancestor relationships between two given names.

In general, the statistics of all datasets after the augmentation methods we discussed in the previous sections are shown in Table~\ref{dataset}.
\begin{table*}[ht]
    \centering
    %\small
    \caption{Dataset statistics after augmentation.}
    \renewcommand{\arraystretch}{0.67} 
    \label{dataset}
    \begin{tabular}{llc}
        \hline
        Dataset & Domain & Size \\ 
        \hline
        \textsc{Kinship} & Parent & 93k \\
        \textsc{City Transportation} & Traffic connection & 135k \\
        \textsc{Ancestor} & Family & 150k \\
        \hline
    \end{tabular}
\end{table*}

\begin{table*}
	\caption{The comparison of models for in-domain learning, and the comparison of cross-domain and cross-form transferability for neuro-symbolic contrastive learning (Neuro-symbolic CL).}
	\centering
 \small
 \renewcommand{\arraystretch}{0.87} 
		\begin{tabular}{lllc}
  \hline
            Train &Test&Model& Accuracy\\ \hline \hline
	      In-domain &&& \\ \hline
		%Baseline & \\ \hline
		\textsc{[Kin $\wedge$ City]-Logic}&\textsc{[Kin $\wedge$ City]-Logic}& BERT-Base & 0.54\\
  \textsc{[Kin $\wedge$ City]-Logic}&\textsc{[Kin $\wedge$ City]-Logic}& Roberta-Base & 0.62\\
  		\textsc{[Kin $\wedge$ City]-Logic}&\textsc{[Kin $\wedge$ City]-Logic}& BERT-Base Neuro-symbolic CL & 0.70\\
            \textsc{[Kin $\wedge$ City]-Logic}&\textsc{[Kin $\wedge$ City]-Logic}&Roberta-Base Neuro-symbolic CL & \textbf{0.74}\\ \hline\hline
            Cross-domain Transfer &&&  \\ \hline
		\textsc{[Kin $\wedge$ City]-Logic}&\textsc{Ancestor-Logic}& BERT-Base & 0.49 \\
  \textsc{[Kin $\wedge$ City]-Logic}&\textsc{Ancestor-Logic}& Roberta-Base & 0.45 \\
            \textsc{[Kin $\wedge$ City]-Logic} & \textsc{Ancestor-Logic} &BERT-Base Neuro-symbolic CL & 0.63\\
                        		\textsc{[Kin $\wedge$ City]-Logic}&\textsc{Ancestor-Logic}& Roberta-Base Neuro-symbolic CL & \textbf{0.64} \\
                          \hline   \hline
            Cross-form Transfer &&&  \\ \hline
		\textsc{[Kin $\wedge$ City]-Logic} & \textsc{[Kin $\wedge$ City]-NL}& BERT-Base & 0.51 \\
  \textsc{[Kin $\wedge$ City]-Logic} & \textsc{[Kin $\wedge$ City]-NL}& Roberta-Base & 0.53 \\
            \textsc{[Kin $\wedge$ City]-Logic} & \textsc{[Kin $\wedge$ City]-NL}& BERT-Base Neuro-symbolic CL & 0.58\\
            		\textsc{[Kin $\wedge$ City]-Logic} & \textsc{[Kin $\wedge$ City]-NL}& Roberta-Base Neuro-symbolic CL & \textbf{0.62} \\\hline  
		\end{tabular}%

 \label{cross}
\end{table*}

\subsection{Result}
\subsubsection{Natural Language vs. Logical Form
Expressions for NLI}\label{preliminary}
%\subsubsection{Natural Language V.S. Logic Form Expressions for NLI}
% To answer the following question: \textit{Is natural language the best form to evaluate performance for logic-based reasoning tasks?}, we evaluate the performance of \textbf{SBERT-SimCSE} model with batch size 16, and the
% maximum text length of the encoder is 512. Results are shown in Table 1:

With the inherent challenge of directly modeling the topological space of logical functions using a Pre-trained Language Model (PLM), we steer our focus towards mapping natural language to logical rules, a relatively straightforward task for natural language processing. %Through our experimentation, we aim to exhibit that Spearman's correlation remains robust when comparing both datasets formed from natural language and logical form.
Our first experiment explores the performance of natural language compares with logical form expressions using our constructed logic-based dataset.

We subject existing sentence embedding methods to evaluate the difference between logic form and natural language form. The evaluation made use of the BERT-base model, fine-tuned on both natural language and logical form datasets. Settings for this experiment included a batch size of 16 and a maximum text length set to 512 for the encoder. 

To evaluate the models, we use Spearman's correlation complemented with accuracy metrics. Spearman's correlation is a rank correlation method that does not assume a linear relationship, making it suitable for our task. By using both Spearman's correlation and accuracy, we can ensure comprehensive evaluation: while accuracy provides a direct measure of correct predictions, the correlation gives an indication in terms of the relationships between data points.

\begin{table}[H]
    \centering
    \renewcommand{\arraystretch}{0.48} % Adjust this value to decrease or increase row spacing
    \small
    \caption{The comparison of data representation on single and multiple domains dataset.  L and NL denote logical form and natural language.}
    \label{comparison}
    \begin{tabular}{lcc}
        \hline
        Dataset & Spearman's correlation & Accuracy \\ 
        \hline\hline
        \textsc{Kinship-L} & 0.69 & 0.63 \\
        \textsc{Kinship-NL} & 0.59 & 0.59 \\
        \hline
        \textsc{City Trans-L} & 0.55 & 0.60 \\
        \textsc{City Trans-NL} & 0.31 & 0.52 \\
        \hline
        \textsc{[Kin $\wedge$ City]-L} & 0.49 & 0.54 \\
        \textsc{[Kin $\wedge$ City]-NL} & 0.39 & 0.48 \\
        \hline 
    \end{tabular}
\end{table}

As shown in Table~\ref{comparison}, after changing the logical form to natural language on \textsc{Kinship} dataset, the Spearman's correlation drops from $0.69$ to $0.59$. This indicates that language models can learn from logic form better on the logic reasoning task (sparse task) we proposed. This can also be confirmed on the \textsc{City Trans} and \textsc{[Kin $\wedge$ City]} datasets.

%(2) we then explore the cross-domain performances of existing sentence embedding methods by first training the model on \textsc{[Kin $\wedge$ City]-Logic} datasets and test on \textsc{Ancestor-Logic} dataset, the poor result indicates that the existing training approach struggles to generalize the learned reasoning logic to a different domain, due to the reliance on shallow heuristics;%\footnote{\todo{SLU: ``The poor result\dots'' should be the start of a new sentence.}} 
%(3) finally we check the cross-form performances by training on the logical form and testing on the natural language form on the \textsc{[Kin $\wedge$ City]} dataset, even the logical form shares the same logical meaning with the natural language form, it proves to be difficult to transfer learned knowledge between two forms.%\footnote{\todo{SLU: Start a new sentence at ``Even the logical form\dots'', and I think there needs to be a ``when'' in there: ``Even when the logical form\dots''}}
%In general, our results show that existing models trained on natural language have the deficiency of relying on shallow heuristics to guess the correct label, leading to the significance of considering logic information in the training loss (our proposed neuro-symbolic contrastive learning loss).

% The comparison of data representation on single and multiple domains dataset (Spearman’s correlation). 
\subsubsection{Neuro-Symbolic Contrastive Learning for Cross-Domain Logic Reasoning}
%In previous experiments, we have verified the effectiveness of using logic-based information for natural language inference tasks. 
We follow the training paradigm of the baseline model in the previous section but use our proposed contrastive learning loss (Equation~\ref{clloss}) and explore the performance of our proposed methods on in-domain, cross-domain, and cross-form scenarios. 
As shown in Table~\ref{cross}, we find that while both BERT-base and Roberta-base models present the poor performance of the baseline training approach on domain transfer tasks, our proposed neuro-symbolic contrastive learning framework can serve as a powerful way to improve the transferability. For both cross-domain transfer and cross-form transfer, our method performs better in overcoming the accuracy drop according to the baseline training approaches, and makes competitive performance even compared with in-domain scenarios.
%This result emphasizes the robustness of 

%Our approach of dataset construction heavily relies on ILP datasets. However, the source of ILP datasets is limited and most ILP datasets are extremely small-scale. Also, our method for pairing hard examples requires that similar underlying logic meta-rules can always be found within given datasets from distinct domains. Due to these shortages, the gap between the source of symbolic and natural language reasoning datasets makes it hard to apply the framework we proposed to more challenging and various tasks.
%修改
%Our dataset construction approach heavily relies on ILP datasets. However, the source of ILP datasets is limited and most of the ILP datasets are extremely small-scale. Additionally, our method for pairing hard examples requires that similar underlying logic meta-rules can always be found within given datasets among distinct domains. Due to these shortages, the gap between the source of symbolic and natural language reasoning datasets makes it hard to apply the framework we proposed to a wider range of complex tasks and further ablation studies. To improve this weakness of our method, a potential future direction is using LLMs to translate existing natural language datasets into well-formed logical arguments and then measure the distance in terms of logical similarity\cite{amgoud2018measuring} with the anchor example.

%\section{Discussion and Conclusion}
\section{Conclusion}

This paper introduces a neuro-symbolic contrastive learning framework that integrates Inductive Logic Programming (ILP) with neural networks to enhance logical reasoning in natural language inference tasks. The framework aims to minimise the distance \(d(E, E^+)\) to enforce logical consistency and maximise \(d(E, E^-)\) to capitalise on logical deviations, thereby refining the model's capacity to discern fine-grained logical distinctions in the embedding space. 

Experimental results demonstrate that our data augmentation method significantly enhances logic inference performance in both natural language and symbolic forms. Additionally, multi-domain fine-tuning within our framework improves the transferability of pre-trained language models across various domains. Our empirical findings align with and extend the assumptions of \cite{shen2022textual} regarding Textual Enhanced Contrastive Learning for solving math word problems, though our approach uniquely incorporates ILP for rule-guided analysis and evaluate on both logic-form and NL-form, adding a novel dimension to the methodology.

The integration of symbolic logic rules and their natural language representations with neural network methodologies not only significantly improves model performance but also underscores the potential for developing deeper, more interpretable architectures for complex reasoning tasks.

\section*{Acknowledgments}

This work was supported by JSPS KAKENHI Grant Number JP21H04905 and JST CREST Grant Number JPMJCR22D3.

\nocite{*}
\bibliographystyle{eptcs}
\bibliography{custom}

\begin{thebibliography}{10}
\providecommand{\bibitemdeclare}[2]{}
\providecommand{\surnamestart}{}
\providecommand{\surnameend}{}
\providecommand{\urlprefix}{Available at }
\providecommand{\url}[1]{\texttt{#1}}
\providecommand{\href}[2]{\texttt{#2}}
\providecommand{\urlalt}[2]{\href{#1}{#2}}
\providecommand{\doi}[1]{doi:\urlalt{https://doi.org/#1}{#1}}
\providecommand{\eprint}[1]{arXiv:\urlalt{https://arxiv.org/abs/#1}{#1}}
\providecommand{\bibinfo}[2]{#2}

\bibitemdeclare{article}{Ando2005}
\bibitem{Ando2005}
\bibinfo{author}{Rie~Kubota \surnamestart Ando\surnameend} \&
  \bibinfo{author}{Tong \surnamestart Zhang\surnameend} (\bibinfo{year}{2005}):
  \emph{\bibinfo{title}{A Framework for Learning Predictive Structures from
  Multiple Tasks and Unlabeled Data}}.
\newblock {\slshape \bibinfo{journal}{Journal of Machine Learning Research}}
  \bibinfo{volume}{6}, pp. \bibinfo{pages}{1817--1853},
  \doi{10.5555/1046920.1194905}.

\bibitemdeclare{inproceedings}{andrew2007scalable}
\bibitem{andrew2007scalable}
\bibinfo{author}{Galen \surnamestart Andrew\surnameend} \&
  \bibinfo{author}{Jianfeng \surnamestart Gao\surnameend}
  (\bibinfo{year}{2007}): \emph{\bibinfo{title}{Scalable training of
  {$L_1$}-regularized log-linear models}}.
\newblock In: {\slshape \bibinfo{booktitle}{Proceedings of the 24th
  International Conference on Machine Learning}}, pp. \bibinfo{pages}{33--40},
  \doi{10.1145/1273496.1273501}.

\bibitemdeclare{inproceedings}{bos2005recognising}
\bibitem{bos2005recognising}
\bibinfo{author}{Johan \surnamestart Bos\surnameend} \& \bibinfo{author}{Katja
  \surnamestart Markert\surnameend} (\bibinfo{year}{2005}):
  \emph{\bibinfo{title}{Recognising textual entailment with logical
  inference}}.
\newblock In: {\slshape \bibinfo{booktitle}{Proceedings of Human Language
  Technology Conference and Conference on Empirical Methods in Natural Language
  Processing}}, pp. \bibinfo{pages}{628--635}, \doi{10.3115/1220575.1220654}.

\bibitemdeclare{inproceedings}{bowman2019deep}
\bibitem{bowman2019deep}
\bibinfo{author}{Samuel \surnamestart Bowman\surnameend} \&
  \bibinfo{author}{Xiaodan \surnamestart Zhu\surnameend}
  (\bibinfo{year}{2019}): \emph{\bibinfo{title}{Deep learning for natural
  language inference}}.
\newblock In: {\slshape \bibinfo{booktitle}{Proceedings of the 2019 Conference
  of the North American Chapter of the Association for Computational
  Linguistics: Tutorials}}, pp. \bibinfo{pages}{6--8},
  \doi{10.18653/v1/N19-5002}.

\bibitemdeclare{inproceedings}{bowman-etal-2015-large}
\bibitem{bowman-etal-2015-large}
\bibinfo{author}{Samuel~R. \surnamestart Bowman\surnameend},
  \bibinfo{author}{Gabor \surnamestart Angeli\surnameend},
  \bibinfo{author}{Christopher \surnamestart Potts\surnameend} \&
  \bibinfo{author}{Christopher~D. \surnamestart Manning\surnameend}
  (\bibinfo{year}{2015}): \emph{\bibinfo{title}{A large annotated corpus for
  learning natural language inference}}.
\newblock In: {\slshape \bibinfo{booktitle}{Proceedings of the 2015 Conference
  on Empirical Methods in Natural Language Processing}},
  \bibinfo{publisher}{Association for Computational Linguistics},
  \bibinfo{address}{Lisbon, Portugal}, pp. \bibinfo{pages}{632--642},
  \doi{10.18653/v1/D15-1075}.

\bibitemdeclare{article}{bratko1995applications}
\bibitem{bratko1995applications}
\bibinfo{author}{Ivan \surnamestart Bratko\surnameend} \&
  \bibinfo{author}{Stephen \surnamestart Muggleton\surnameend}
  (\bibinfo{year}{1995}): \emph{\bibinfo{title}{Applications of inductive logic
  programming}}.
\newblock {\slshape \bibinfo{journal}{Communications of the ACM}}
  \bibinfo{volume}{38}(\bibinfo{number}{11}), pp. \bibinfo{pages}{65--70},
  \doi{10.1145/219717.219771}.

\bibitemdeclare{inproceedings}{calo2022enhancing}
\bibitem{calo2022enhancing}
\bibinfo{author}{Eduardo \surnamestart Cal{\`o}\surnameend},
  \bibinfo{author}{Elze \surnamestart van~der Werf\surnameend},
  \bibinfo{author}{Albert \surnamestart Gatt\surnameend} \&
  \bibinfo{author}{Kees \surnamestart van Deemter\surnameend}
  (\bibinfo{year}{2022}): \emph{\bibinfo{title}{Enhancing and Evaluating the
  Grammatical Framework Approach to Logic-to-Text Generation}}.
\newblock In: {\slshape \bibinfo{booktitle}{Proceedings of the 2nd Workshop on
  Natural Language Generation, Evaluation, and Metrics (GEM)}}, pp.
  \bibinfo{pages}{148--171}, \doi{10.18653/v1/2022.gem-1.13}.

\bibitemdeclare{inproceedings}{chen2020simple}
\bibitem{chen2020simple}
\bibinfo{author}{Ting \surnamestart Chen\surnameend}, \bibinfo{author}{Simon
  \surnamestart Kornblith\surnameend}, \bibinfo{author}{Mohammad \surnamestart
  Norouzi\surnameend} \& \bibinfo{author}{Geoffrey \surnamestart
  Hinton\surnameend} (\bibinfo{year}{2020}): \emph{\bibinfo{title}{A simple
  framework for contrastive learning of visual representations}}.
\newblock In: {\slshape \bibinfo{booktitle}{International conference on machine
  learning}}, \bibinfo{organization}{PMLR}, pp. \bibinfo{pages}{1597--1607},
  \doi{10.48550/arXiv.2002.05709}.

\bibitemdeclare{article}{ct1965}
\bibitem{ct1965}
\bibinfo{author}{James~W. \surnamestart Cooley\surnameend} \&
  \bibinfo{author}{John~W. \surnamestart Tukey\surnameend}
  (\bibinfo{year}{1965}): \emph{\bibinfo{title}{An algorithm for the machine
  calculation of complex {F}ourier series}}.
\newblock {\slshape \bibinfo{journal}{Mathematics of Computation}}
  \bibinfo{volume}{19}(\bibinfo{number}{90}), pp. \bibinfo{pages}{297--301},
  \doi{10.1090/S0025-5718-1965-0178586-1}.

\bibitemdeclare{article}{cropper2022inductive}
\bibitem{cropper2022inductive}
\bibinfo{author}{Andrew \surnamestart Cropper\surnameend} \&
  \bibinfo{author}{Sebastijan \surnamestart Duman\v{c}i\'{c}\surnameend}
  (\bibinfo{year}{2022}): \emph{\bibinfo{title}{Inductive Logic Programming At
  30: A New Introduction}}.
\newblock {\slshape \bibinfo{journal}{J. Artif. Int. Res.}}
  \bibinfo{volume}{74}, \doi{10.1613/jair.1.13507}.

\bibitemdeclare{article}{cropper2021learning}
\bibitem{cropper2021learning}
\bibinfo{author}{Andrew \surnamestart Cropper\surnameend} \&
  \bibinfo{author}{Rolf \surnamestart Morel\surnameend} (\bibinfo{year}{2021}):
  \emph{\bibinfo{title}{Learning programs by learning from failures}}.
\newblock {\slshape \bibinfo{journal}{Machine Learning}} \bibinfo{volume}{110},
  pp. \bibinfo{pages}{801--856}, \doi{10.1007/s10994-020-05934-z}.

\bibitemdeclare{article}{cunnington2023ffnsl}
\bibitem{cunnington2023ffnsl}
\bibinfo{author}{Daniel \surnamestart Cunnington\surnameend},
  \bibinfo{author}{Mark \surnamestart Law\surnameend}, \bibinfo{author}{Jorge
  \surnamestart Lobo\surnameend} \& \bibinfo{author}{Alessandra \surnamestart
  Russo\surnameend} (\bibinfo{year}{2023}): \emph{\bibinfo{title}{Ffnsl:
  Feed-forward neural-symbolic learner}}.
\newblock {\slshape \bibinfo{journal}{Machine Learning}}
  \bibinfo{volume}{112}(\bibinfo{number}{2}), pp. \bibinfo{pages}{515--569},
  \doi{10.1007/s10994-022-06278-6}.

\bibitemdeclare{inproceedings}{cunnington2022neuro}
\bibitem{cunnington2022neuro}
\bibinfo{author}{Daniel \surnamestart Cunnington\surnameend},
  \bibinfo{author}{Mark \surnamestart Law\surnameend}, \bibinfo{author}{Jorge
  \surnamestart Lobo\surnameend} \& \bibinfo{author}{Alessandra \surnamestart
  Russo\surnameend} (\bibinfo{year}{2023}):
  \emph{\bibinfo{title}{Neuro-symbolic learning of answer set programs from raw
  data}}.
\newblock In: {\slshape \bibinfo{booktitle}{Proceedings of the Thirty-Second
  International Joint Conference on Artificial Intelligence}},
  \bibinfo{series}{IJCAI '23}, \doi{10.24963/ijcai.2023/399}.

\bibitemdeclare{article}{dong2019neural}
\bibitem{dong2019neural}
\bibinfo{author}{Honghua \surnamestart Dong\surnameend},
  \bibinfo{author}{Jiayuan \surnamestart Mao\surnameend}, \bibinfo{author}{Tian
  \surnamestart Lin\surnameend}, \bibinfo{author}{Chong \surnamestart
  Wang\surnameend}, \bibinfo{author}{Lihong \surnamestart Li\surnameend} \&
  \bibinfo{author}{Denny \surnamestart Zhou\surnameend} (\bibinfo{year}{2019}):
  \emph{\bibinfo{title}{Neural logic machines}}.
\newblock {\slshape \bibinfo{journal}{arXiv preprint arXiv:1904.11694}},
  \doi{10.48550/arXiv.1904.11694}.

\bibitemdeclare{article}{evans2018learning}
\bibitem{evans2018learning}
\bibinfo{author}{Richard \surnamestart Evans\surnameend} \&
  \bibinfo{author}{Edward \surnamestart Grefenstette\surnameend}
  (\bibinfo{year}{2018}): \emph{\bibinfo{title}{Learning explanatory rules from
  noisy data}}.
\newblock {\slshape \bibinfo{journal}{Journal of Artificial Intelligence
  Research}} \bibinfo{volume}{61}, pp. \bibinfo{pages}{1--64},
  \doi{10.1613/jair.5714}.

\bibitemdeclare{article}{tacl}
\bibitem{tacl}
\bibinfo{author}{Yufei \surnamestart Feng\surnameend}, \bibinfo{author}{Xiaoyu
  \surnamestart Yang\surnameend}, \bibinfo{author}{Xiaodan \surnamestart
  Zhu\surnameend} \& \bibinfo{author}{Michael \surnamestart
  Greenspan\surnameend} (\bibinfo{year}{2022}):
  \emph{\bibinfo{title}{{Neuro-symbolic Natural Logic with Introspective
  Revision for Natural Language Inference}}}.
\newblock {\slshape \bibinfo{journal}{Transactions of the Association for
  Computational Linguistics}} \bibinfo{volume}{10}, pp.
  \bibinfo{pages}{240--256}, \doi{10.1162/tacl_a_00458}.

\bibitemdeclare{inproceedings}{feng2020exploring}
\bibitem{feng2020exploring}
\bibinfo{author}{Yufei \surnamestart Feng\surnameend}, \bibinfo{author}{Zi{'}ou
  \surnamestart Zheng\surnameend}, \bibinfo{author}{Quan \surnamestart
  Liu\surnameend}, \bibinfo{author}{Michael \surnamestart Greenspan\surnameend}
  \& \bibinfo{author}{Xiaodan \surnamestart Zhu\surnameend}
  (\bibinfo{year}{2020}): \emph{\bibinfo{title}{Exploring End-to-End
  Differentiable Natural Logic Modeling}}.
\newblock In: {\slshape \bibinfo{booktitle}{Proceedings of the 28th
  International Conference on Computational Linguistics}},
  \bibinfo{publisher}{International Committee on Computational Linguistics},
  pp. \bibinfo{pages}{1172--1185}, \doi{10.18653/v1/2020.coling-main.101}.

\bibitemdeclare{inproceedings}{gao2021simcse}
\bibitem{gao2021simcse}
\bibinfo{author}{Tianyu \surnamestart Gao\surnameend},
  \bibinfo{author}{Xingcheng \surnamestart Yao\surnameend} \&
  \bibinfo{author}{Danqi \surnamestart Chen\surnameend} (\bibinfo{year}{2021}):
  \emph{\bibinfo{title}{{S}im{CSE}: Simple Contrastive Learning of Sentence
  Embeddings}}.
\newblock In: {\slshape \bibinfo{booktitle}{Proceedings of the 2021 Conference
  on Empirical Methods in Natural Language Processing}},
  \bibinfo{publisher}{Association for Computational Linguistics}, pp.
  \bibinfo{pages}{6894--6910}, \doi{10.18653/v1/2021.emnlp-main.552}.

\bibitemdeclare{inproceedings}{gururangan2018annotation}
\bibitem{gururangan2018annotation}
\bibinfo{author}{Suchin \surnamestart Gururangan\surnameend},
  \bibinfo{author}{Swabha \surnamestart Swayamdipta\surnameend},
  \bibinfo{author}{Omer \surnamestart Levy\surnameend}, \bibinfo{author}{Roy
  \surnamestart Schwartz\surnameend}, \bibinfo{author}{Samuel \surnamestart
  Bowman\surnameend} \& \bibinfo{author}{Noah~A. \surnamestart
  Smith\surnameend} (\bibinfo{year}{2018}): \emph{\bibinfo{title}{Annotation
  Artifacts in Natural Language Inference Data}}.
\newblock In: {\slshape \bibinfo{booktitle}{Proceedings of the 2018 Conference
  of the North {A}merican Chapter of the Association for Computational
  Linguistics: Human Language Technologies, Volume 2 (Short Papers)}}, pp.
  \bibinfo{pages}{107--112}, \doi{10.18653/v1/N18-2017}.

\bibitemdeclare{book}{Gusfield:97}
\bibitem{Gusfield:97}
\bibinfo{author}{Dan \surnamestart Gusfield\surnameend} (\bibinfo{year}{1997}):
  \emph{\bibinfo{title}{Algorithms on Strings, Trees and Sequences}}.
\newblock \bibinfo{publisher}{Cambridge University Press},
  \bibinfo{address}{Cambridge, UK}, \doi{10.1017/CBO9780511574931}.

\bibitemdeclare{article}{hamilton2022neuro}
\bibitem{hamilton2022neuro}
\bibinfo{author}{Kyle \surnamestart Hamilton\surnameend},
  \bibinfo{author}{Aparna \surnamestart Nayak\surnameend},
  \bibinfo{author}{Bojan \surnamestart Božić\surnameend} \&
  \bibinfo{author}{Luca \surnamestart Longo\surnameend} (\bibinfo{year}{2022}):
  \emph{\bibinfo{title}{Is neuro-symbolic AI meeting its promises in natural
  language processing? A structured review}}.
\newblock {\slshape \bibinfo{journal}{Semantic Web}}, p.
  \bibinfo{pages}{1–42}, \doi{10.3233/sw-223228}.

\bibitemdeclare{inproceedings}{9157636}
\bibitem{9157636}
\bibinfo{author}{Kaiming \surnamestart He\surnameend}, \bibinfo{author}{Haoqi
  \surnamestart Fan\surnameend}, \bibinfo{author}{Yuxin \surnamestart
  Wu\surnameend}, \bibinfo{author}{Saining \surnamestart Xie\surnameend} \&
  \bibinfo{author}{Ross \surnamestart Girshick\surnameend}
  (\bibinfo{year}{2020}): \emph{\bibinfo{title}{Momentum Contrast for
  Unsupervised Visual Representation Learning}}.
\newblock In: {\slshape \bibinfo{booktitle}{2020 IEEE/CVF Conference on
  Computer Vision and Pattern Recognition (CVPR)}}, pp.
  \bibinfo{pages}{9726--9735}, \doi{10.1109/CVPR42600.2020.00975}.

\bibitemdeclare{article}{hochreiter1997long}
\bibitem{hochreiter1997long}
\bibinfo{author}{Sepp \surnamestart Hochreiter\surnameend} \&
  \bibinfo{author}{J{\"u}rgen \surnamestart Schmidhuber\surnameend}
  (\bibinfo{year}{1997}): \emph{\bibinfo{title}{Long short-term memory}}.
\newblock {\slshape \bibinfo{journal}{Neural computation}}
  \bibinfo{volume}{9}(\bibinfo{number}{8}), pp. \bibinfo{pages}{1735--1780},
  \doi{10.1162/neco.1997.9.8.1735}.

\bibitemdeclare{article}{kaminski2018exploiting}
\bibitem{kaminski2018exploiting}
\bibinfo{author}{Tobias \surnamestart Kaminski\surnameend},
  \bibinfo{author}{Thomas \surnamestart Eiter\surnameend} \&
  \bibinfo{author}{Katsumi \surnamestart Inoue\surnameend}
  (\bibinfo{year}{2018}): \emph{\bibinfo{title}{Exploiting answer set
  programming with external sources for meta-interpretive learning}}.
\newblock {\slshape \bibinfo{journal}{Theory and Practice of Logic
  Programming}} \bibinfo{volume}{18}(\bibinfo{number}{3-4}), pp.
  \bibinfo{pages}{571--588}, \doi{10.1017/S1471068418000261}.

\bibitemdeclare{article}{kautz2022third}
\bibitem{kautz2022third}
\bibinfo{author}{Henry \surnamestart Kautz\surnameend} (\bibinfo{year}{2022}):
  \emph{\bibinfo{title}{The third ai summer: Aaai robert s. engelmore memorial
  lecture}}.
\newblock {\slshape \bibinfo{journal}{AI Magazine}}
  \bibinfo{volume}{43}(\bibinfo{number}{1}), pp. \bibinfo{pages}{105--125},
  \doi{10.1002/aaai.12036}.

\bibitemdeclare{article}{le2020contrastive}
\bibitem{le2020contrastive}
\bibinfo{author}{Phuc~H. \surnamestart Le-Khac\surnameend},
  \bibinfo{author}{Graham \surnamestart Healy\surnameend} \&
  \bibinfo{author}{Alan~F. \surnamestart Smeaton\surnameend}
  (\bibinfo{year}{2020}): \emph{\bibinfo{title}{Contrastive Representation
  Learning: A Framework and Review}}.
\newblock {\slshape \bibinfo{journal}{IEEE Access}} \bibinfo{volume}{8}, pp.
  \bibinfo{pages}{193907--193934}, \doi{10.1109/ACCESS.2020.3031549}.

\bibitemdeclare{inproceedings}{luo2022simple}
\bibitem{luo2022simple}
\bibinfo{author}{Cheng \surnamestart Luo\surnameend}, \bibinfo{author}{Wei
  \surnamestart Liu\surnameend}, \bibinfo{author}{Jieyu \surnamestart
  Lin\surnameend}, \bibinfo{author}{Jiajie \surnamestart Zou\surnameend},
  \bibinfo{author}{Ming \surnamestart Xiang\surnameend} \& \bibinfo{author}{Nai
  \surnamestart Ding\surnameend} (\bibinfo{year}{2022}):
  \emph{\bibinfo{title}{Simple but Challenging: Natural Language Inference
  Models Fail on Simple Sentences}}.
\newblock In \bibinfo{editor}{Yoav \surnamestart Goldberg\surnameend},
  \bibinfo{editor}{Zornitsa \surnamestart Kozareva\surnameend} \&
  \bibinfo{editor}{Yue \surnamestart Zhang\surnameend}, editors: {\slshape
  \bibinfo{booktitle}{Findings of the Association for Computational
  Linguistics: EMNLP 2022}}, \bibinfo{publisher}{Association for Computational
  Linguistics}, pp. \bibinfo{pages}{3449--3462},
  \doi{10.18653/v1/2022.findings-emnlp.252}.

\bibitemdeclare{inproceedings}{manhaeve2018deepproblog}
\bibitem{manhaeve2018deepproblog}
\bibinfo{author}{Robin \surnamestart Manhaeve\surnameend},
  \bibinfo{author}{Sebastijan \surnamestart Dumancic\surnameend},
  \bibinfo{author}{Angelika \surnamestart Kimmig\surnameend},
  \bibinfo{author}{Thomas \surnamestart Demeester\surnameend} \&
  \bibinfo{author}{Luc~De \surnamestart Raedt\surnameend}
  (\bibinfo{year}{2018}): \emph{\bibinfo{title}{DeepProbLog: neural
  probabilistic logic programming}}.
\newblock In: {\slshape \bibinfo{booktitle}{Proceedings of the 32nd
  International Conference on Neural Information Processing Systems}},
  \bibinfo{series}{NIPS'18}, p. \bibinfo{pages}{3753–3763},
  \doi{10.5555/3327144.3327291}.

\bibitemdeclare{article}{mao2019neuro}
\bibitem{mao2019neuro}
\bibinfo{author}{Jiayuan \surnamestart Mao\surnameend}, \bibinfo{author}{Chuang
  \surnamestart Gan\surnameend}, \bibinfo{author}{Pushmeet \surnamestart
  Kohli\surnameend}, \bibinfo{author}{Joshua~B. \surnamestart
  Tenenbaum\surnameend} \& \bibinfo{author}{Jiajun \surnamestart Wu\surnameend}
  (\bibinfo{year}{2019}): \emph{\bibinfo{title}{The Neuro-Symbolic Concept
  Learner: Interpreting Scenes, Words, and Sentences From Natural
  Supervision}}.
\newblock {\slshape \bibinfo{journal}{arXiv preprint arXiv:1904.12584}},
  \doi{10.48550/arXiv.1904.12584}.

\bibitemdeclare{inproceedings}{mccoy2019right}
\bibitem{mccoy2019right}
\bibinfo{author}{Tom \surnamestart McCoy\surnameend}, \bibinfo{author}{Ellie
  \surnamestart Pavlick\surnameend} \& \bibinfo{author}{Tal \surnamestart
  Linzen\surnameend} (\bibinfo{year}{2019}): \emph{\bibinfo{title}{Right for
  the Wrong Reasons: Diagnosing Syntactic Heuristics in Natural Language
  Inference}}.
\newblock In: {\slshape \bibinfo{booktitle}{Proceedings of the 57th Annual
  Meeting of the Association for Computational Linguistics}}, pp.
  \bibinfo{pages}{3428--3448}, \doi{10.18653/v1/P19-1334}.

\bibitemdeclare{inproceedings}{mitra2019declarative}
\bibitem{mitra2019declarative}
\bibinfo{author}{Arindam \surnamestart Mitra\surnameend},
  \bibinfo{author}{Peter \surnamestart Clark\surnameend},
  \bibinfo{author}{Oyvind \surnamestart Tafjord\surnameend} \&
  \bibinfo{author}{Chitta \surnamestart Baral\surnameend}
  (\bibinfo{year}{2019}): \emph{\bibinfo{title}{Declarative question answering
  over knowledge bases containing natural language text with answer set
  programming}}.
\newblock In: {\slshape \bibinfo{booktitle}{Proceedings of the AAAI Conference
  on Artificial Intelligence}}, \bibinfo{volume}{01}, pp.
  \bibinfo{pages}{3003--3010}, \doi{10.1609/aaai.v33i01.33013003}.

\bibitemdeclare{article}{muggleton1992inductive}
\bibitem{muggleton1992inductive}
\bibinfo{author}{Stephen \surnamestart Muggleton\surnameend}
  (\bibinfo{year}{1991}): \emph{\bibinfo{title}{Inductive logic programming}}.
\newblock {\slshape \bibinfo{journal}{New generation computing}}
  \bibinfo{volume}{8}, pp. \bibinfo{pages}{295--318}, \doi{10.1007/BF03037089}.

\bibitemdeclare{article}{muggleton1994inductive}
\bibitem{muggleton1994inductive}
\bibinfo{author}{Stephen \surnamestart Muggleton\surnameend} \&
  \bibinfo{author}{Luc \surnamestart De~Raedt\surnameend}
  (\bibinfo{year}{1994}): \emph{\bibinfo{title}{Inductive logic programming:
  Theory and methods}}.
\newblock {\slshape \bibinfo{journal}{The Journal of Logic Programming}}
  \bibinfo{volume}{19}, pp. \bibinfo{pages}{629--679},
  \doi{10.1016/0743-1066(94)90035-3}.

\bibitemdeclare{article}{muggleton2012ilp}
\bibitem{muggleton2012ilp}
\bibinfo{author}{Stephen \surnamestart Muggleton\surnameend},
  \bibinfo{author}{Luc \surnamestart De~Raedt\surnameend},
  \bibinfo{author}{David \surnamestart Poole\surnameend}, \bibinfo{author}{Ivan
  \surnamestart Bratko\surnameend}, \bibinfo{author}{Peter \surnamestart
  Flach\surnameend}, \bibinfo{author}{Katsumi \surnamestart Inoue\surnameend}
  \& \bibinfo{author}{Ashwin \surnamestart Srinivasan\surnameend}
  (\bibinfo{year}{2012}): \emph{\bibinfo{title}{ILP turns 20: biography and
  future challenges}}.
\newblock {\slshape \bibinfo{journal}{Machine learning}} \bibinfo{volume}{86},
  pp. \bibinfo{pages}{3--23}, \doi{10.1007/s10994-011-5259-2}.

\bibitemdeclare{inproceedings}{nangia2019human}
\bibitem{nangia2019human}
\bibinfo{author}{Nikita \surnamestart Nangia\surnameend} \&
  \bibinfo{author}{Samuel~R. \surnamestart Bowman\surnameend}
  (\bibinfo{year}{2019}): \emph{\bibinfo{title}{Human vs. Muppet: A
  Conservative Estimate of Human Performance on the {GLUE} Benchmark}}.
\newblock In: {\slshape \bibinfo{booktitle}{Proceedings of the 57th Annual
  Meeting of the Association for Computational Linguistics}}, pp.
  \bibinfo{pages}{4566--4575}, \doi{10.18653/v1/P19-1449}.

\bibitemdeclare{inproceedings}{oh2016deep}
\bibitem{oh2016deep}
\bibinfo{author}{Hyun \surnamestart Oh~Song\surnameend},
  \bibinfo{author}{Yu~\surnamestart Xiang\surnameend},
  \bibinfo{author}{Stefanie \surnamestart Jegelka\surnameend} \&
  \bibinfo{author}{Silvio \surnamestart Savarese\surnameend}
  (\bibinfo{year}{2016}): \emph{\bibinfo{title}{Deep metric learning via lifted
  structured feature embedding}}.
\newblock In: {\slshape \bibinfo{booktitle}{Proceedings of the IEEE conference
  on computer vision and pattern recognition}}, pp.
  \bibinfo{pages}{4004--4012}, \doi{10.1109/CVPR.2016.434}.

\bibitemdeclare{article}{payani2019inductive}
\bibitem{payani2019inductive}
\bibinfo{author}{Ali \surnamestart Payani\surnameend} \&
  \bibinfo{author}{Faramarz \surnamestart Fekri\surnameend}
  (\bibinfo{year}{2019}): \emph{\bibinfo{title}{Inductive logic programming via
  differentiable deep neural logic networks}}.
\newblock {\slshape \bibinfo{journal}{arXiv preprint arXiv:1906.03523}},
  \doi{10.48550/arXiv.1906.03523}.

\bibitemdeclare{article}{pendharkar2022asp}
\bibitem{pendharkar2022asp}
\bibinfo{author}{Dhruva \surnamestart Pendharkar\surnameend},
  \bibinfo{author}{Kinjal \surnamestart Basu\surnameend},
  \bibinfo{author}{Farhad \surnamestart Shakerin\surnameend} \&
  \bibinfo{author}{Gopal \surnamestart Gupta\surnameend}
  (\bibinfo{year}{2022}): \emph{\bibinfo{title}{An asp-based approach to
  answering natural language questions for texts}}.
\newblock {\slshape \bibinfo{journal}{Theory and Practice of Logic
  Programming}} \bibinfo{volume}{22}(\bibinfo{number}{3}), pp.
  \bibinfo{pages}{419--443}, \doi{10.1007/978-3-030-05998-9_4}.

\bibitemdeclare{article}{ranta2004grammatical}
\bibitem{ranta2004grammatical}
\bibinfo{author}{Aarne \surnamestart Ranta\surnameend} (\bibinfo{year}{2004}):
  \emph{\bibinfo{title}{Grammatical framework}}.
\newblock {\slshape \bibinfo{journal}{Journal of Functional Programming}}
  \bibinfo{volume}{14}(\bibinfo{number}{2}), pp. \bibinfo{pages}{145--189},
  \doi{10.1017/S0956796803004738}.

\bibitemdeclare{article}{rasooli-tetrault-2015}
\bibitem{rasooli-tetrault-2015}
\bibinfo{author}{Mohammad~Sadegh \surnamestart Rasooli\surnameend} \&
  \bibinfo{author}{Joel~R. \surnamestart Tetreault\surnameend}
  (\bibinfo{year}{2015}): \emph{\bibinfo{title}{Yara Parser: {A} Fast and
  Accurate Dependency Parser}}.
\newblock {\slshape \bibinfo{journal}{arXiv preprint arXiv:1503.06733v2}},
  \doi{10.48550/arXiv.1503.06733}.

\bibitemdeclare{inproceedings}{reimers-gurevych-2019-sentence}
\bibitem{reimers-gurevych-2019-sentence}
\bibinfo{author}{Nils \surnamestart Reimers\surnameend} \&
  \bibinfo{author}{Iryna \surnamestart Gurevych\surnameend}
  (\bibinfo{year}{2019}): \emph{\bibinfo{title}{Sentence-{BERT}: Sentence
  Embeddings using {S}iamese {BERT}-Networks}}.
\newblock In: {\slshape \bibinfo{booktitle}{Proceedings of the 2019 Conference
  on Empirical Methods in Natural Language Processing and the 9th International
  Joint Conference on Natural Language Processing}}, pp.
  \bibinfo{pages}{3982--3992}, \doi{10.18653/v1/D19-1410}.

\bibitemdeclare{incollection}{reiter1981closed}
\bibitem{reiter1981closed}
\bibinfo{author}{Raymond \surnamestart Reiter\surnameend}
  (\bibinfo{year}{1981}): \emph{\bibinfo{title}{On closed world data bases}}.
\newblock In: {\slshape \bibinfo{booktitle}{Readings in artificial
  intelligence}}, \bibinfo{publisher}{Elsevier}, pp. \bibinfo{pages}{119--140},
  \doi{10.1016/B978-0-934613-03-3.50014-3}.

\bibitemdeclare{inproceedings}{rosenman2020exposing}
\bibitem{rosenman2020exposing}
\bibinfo{author}{Shachar \surnamestart Rosenman\surnameend},
  \bibinfo{author}{Alon \surnamestart Jacovi\surnameend} \&
  \bibinfo{author}{Yoav \surnamestart Goldberg\surnameend}
  (\bibinfo{year}{2020}): \emph{\bibinfo{title}{{E}xposing {S}hallow
  {H}euristics of {R}elation {E}xtraction {M}odels with {C}hallenge {D}ata}}.
\newblock In: {\slshape \bibinfo{booktitle}{Proceedings of the 2020 Conference
  on Empirical Methods in Natural Language Processing}}, pp.
  \bibinfo{pages}{3702--3710}, \doi{10.18653/v1/2020.emnlp-main.302}.

\bibitemdeclare{inproceedings}{sen2022neuro}
\bibitem{sen2022neuro}
\bibinfo{author}{Prithviraj \surnamestart Sen\surnameend},
  \bibinfo{author}{Breno~WSR \surnamestart de~Carvalho\surnameend},
  \bibinfo{author}{Ryan \surnamestart Riegel\surnameend} \&
  \bibinfo{author}{Alexander \surnamestart Gray\surnameend}
  (\bibinfo{year}{2022}): \emph{\bibinfo{title}{Neuro-symbolic inductive logic
  programming with logical neural networks}}.
\newblock In: {\slshape \bibinfo{booktitle}{Proceedings of the AAAI Conference
  on Artificial Intelligence}}, \bibinfo{volume}{36}, pp.
  \bibinfo{pages}{8212--8219}, \doi{10.1609/aaai.v36i8.20795}.

\bibitemdeclare{inproceedings}{shen2022textual}
\bibitem{shen2022textual}
\bibinfo{author}{Yibin \surnamestart Shen\surnameend},
  \bibinfo{author}{Qianying \surnamestart Liu\surnameend},
  \bibinfo{author}{Zhuoyuan \surnamestart Mao\surnameend}, \bibinfo{author}{Fei
  \surnamestart Cheng\surnameend} \& \bibinfo{author}{Sadao \surnamestart
  Kurohashi\surnameend} (\bibinfo{year}{2022}): \emph{\bibinfo{title}{Textual
  Enhanced Contrastive Learning for Solving Math Word Problems}}.
\newblock In: {\slshape \bibinfo{booktitle}{Findings of the Association for
  Computational Linguistics: EMNLP 2022}}, pp. \bibinfo{pages}{4297--4307},
  \doi{10.18653/v1/2022.findings-emnlp.316}.

\bibitemdeclare{inproceedings}{sinha2020unnatural}
\bibitem{sinha2020unnatural}
\bibinfo{author}{Koustuv \surnamestart Sinha\surnameend},
  \bibinfo{author}{Prasanna \surnamestart Parthasarathi\surnameend},
  \bibinfo{author}{Joelle \surnamestart Pineau\surnameend} \&
  \bibinfo{author}{Adina \surnamestart Williams\surnameend}
  (\bibinfo{year}{2021}): \emph{\bibinfo{title}{{UnNatural} {L}anguage
  {I}nference}}.
\newblock In: {\slshape \bibinfo{booktitle}{Proceedings of the 59th Annual
  Meeting of the Association for Computational Linguistics and the 11th
  International Joint Conference on Natural Language Processing (Volume 1: Long
  Papers)}}, pp. \bibinfo{pages}{7329--7346},
  \doi{10.18653/v1/2021.acl-long.569}.

\bibitemdeclare{article}{storks2019recent}
\bibitem{storks2019recent}
\bibinfo{author}{Shane \surnamestart Storks\surnameend},
  \bibinfo{author}{Qiaozi \surnamestart Gao\surnameend} \&
  \bibinfo{author}{Joyce~Y. \surnamestart Chai\surnameend}
  (\bibinfo{year}{2020}): \emph{\bibinfo{title}{Recent Advances in Natural
  Language Inference: A Survey of Benchmarks, Resources, and Approaches}}.
\newblock {\slshape \bibinfo{journal}{arXiv preprint arXiv:1904.11694}},
  \doi{10.48550/arXiv.1904.01172}.

\bibitemdeclare{incollection}{vapnik2015uniform}
\bibitem{vapnik2015uniform}
\bibinfo{author}{Vladimir~N \surnamestart Vapnik\surnameend} \&
  \bibinfo{author}{A~Ya \surnamestart Chervonenkis\surnameend}
  (\bibinfo{year}{2015}): \emph{\bibinfo{title}{On the uniform convergence of
  relative frequencies of events to their probabilities}}.
\newblock In: {\slshape \bibinfo{booktitle}{Measures of complexity: festschrift
  for alexey chervonenkis}}, \bibinfo{publisher}{Springer}, pp.
  \bibinfo{pages}{11--30}, \doi{10.1007/978-3-319-21852-6_3}.

\bibitemdeclare{article}{vaswani2017attention}
\bibitem{vaswani2017attention}
\bibinfo{author}{Ashish \surnamestart Vaswani\surnameend},
  \bibinfo{author}{Noam \surnamestart Shazeer\surnameend},
  \bibinfo{author}{Niki \surnamestart Parmar\surnameend},
  \bibinfo{author}{Jakob \surnamestart Uszkoreit\surnameend},
  \bibinfo{author}{Llion \surnamestart Jones\surnameend},
  \bibinfo{author}{Aidan~N \surnamestart Gomez\surnameend},
  \bibinfo{author}{{\L}ukasz \surnamestart Kaiser\surnameend} \&
  \bibinfo{author}{Illia \surnamestart Polosukhin\surnameend}
  (\bibinfo{year}{2017}): \emph{\bibinfo{title}{Attention is all you need}}.
\newblock {\slshape \bibinfo{journal}{Advances in neural information processing
  systems}} \bibinfo{volume}{30}, \doi{10.48550/arXiv.2002.05709}.

\bibitemdeclare{inproceedings}{williams2017broad}
\bibitem{williams2017broad}
\bibinfo{author}{Adina \surnamestart Williams\surnameend},
  \bibinfo{author}{Nikita \surnamestart Nangia\surnameend} \&
  \bibinfo{author}{Samuel \surnamestart Bowman\surnameend}
  (\bibinfo{year}{2018}): \emph{\bibinfo{title}{A Broad-Coverage Challenge
  Corpus for Sentence Understanding through Inference}}.
\newblock In: {\slshape \bibinfo{booktitle}{Proceedings of the 2018 Conference
  of the North {A}merican Chapter of the Association for Computational
  Linguistics: Human Language Technologies, Volume 1 (Long Papers)}}, pp.
  \bibinfo{pages}{1112--1122}, \doi{10.18653/v1/N18-1101}.

\bibitemdeclare{inproceedings}{williams-etal-2018-broad}
\bibitem{williams-etal-2018-broad}
\bibinfo{author}{Adina \surnamestart Williams\surnameend},
  \bibinfo{author}{Nikita \surnamestart Nangia\surnameend} \&
  \bibinfo{author}{Samuel \surnamestart Bowman\surnameend}
  (\bibinfo{year}{2018}): \emph{\bibinfo{title}{A Broad-Coverage Challenge
  Corpus for Sentence Understanding through Inference}}.
\newblock In: {\slshape \bibinfo{booktitle}{Proceedings of the 2018 Conference
  of the North {A}merican Chapter of the Association for Computational
  Linguistics: Human Language Technologies, Volume 1 (Long Papers)}}, pp.
  \bibinfo{pages}{1112--1122}, \doi{10.18653/v1/N18-1101}.

\bibitemdeclare{inproceedings}{yan2021consert}
\bibitem{yan2021consert}
\bibinfo{author}{Yuanmeng \surnamestart Yan\surnameend}, \bibinfo{author}{Rumei
  \surnamestart Li\surnameend}, \bibinfo{author}{Sirui \surnamestart
  Wang\surnameend}, \bibinfo{author}{Fuzheng \surnamestart Zhang\surnameend},
  \bibinfo{author}{Wei \surnamestart Wu\surnameend} \& \bibinfo{author}{Weiran
  \surnamestart Xu\surnameend} (\bibinfo{year}{2021}):
  \emph{\bibinfo{title}{{C}on{SERT}: A Contrastive Framework for
  Self-Supervised Sentence Representation Transfer}}.
\newblock In: {\slshape \bibinfo{booktitle}{Proceedings of the 59th Annual
  Meeting of the Association for Computational Linguistics and the 11th
  International Joint Conference on Natural Language Processing (Volume 1: Long
  Papers)}}, pp. \bibinfo{pages}{5065--5075},
  \doi{10.18653/v1/2021.acl-long.393}.

\bibitemdeclare{article}{zhang2021neural}
\bibitem{zhang2021neural}
\bibinfo{author}{Jing \surnamestart Zhang\surnameend},
  \bibinfo{author}{Bo~\surnamestart Chen\surnameend}, \bibinfo{author}{Lingxi
  \surnamestart Zhang\surnameend}, \bibinfo{author}{Xirui \surnamestart
  Ke\surnameend} \& \bibinfo{author}{Haipeng \surnamestart Ding\surnameend}
  (\bibinfo{year}{2021}): \emph{\bibinfo{title}{Neural, symbolic and
  neural-symbolic reasoning on knowledge graphs}}.
\newblock {\slshape \bibinfo{journal}{AI Open}} \bibinfo{volume}{2}, pp.
  \bibinfo{pages}{14--35}, \doi{10.1016/j.aiopen.2021.03.001}.

\end{thebibliography}
\end{document}